\documentclass[letterpaper, 10 pt, conference]{ieeeconf}  % Comment this line out if you need a4paper

\IEEEoverridecommandlockouts                              % This command is only needed if 
\makeatletter
\let\NAT@parse\undefined
\makeatother
\usepackage[bookmarks=true,hidelinks]{hyperref}
\usepackage{times}
\usepackage{epsfig}
\usepackage{graphicx}
\usepackage{amsmath, bm}
\usepackage{amssymb}
\usepackage{multirow}
\usepackage[dvipsnames]{xcolor}
\usepackage[acronym]{glossaries}

\title{\LARGE \bf
  Image-Based Place Recognition on Bucolic Environment Across Seasons From Semantic
  Edge Description}

\author{Assia Benbihi$^{1}$, St{\'e}phanie Aravecchia$^{2}$, Matthieu Geist$^{3}$ and  C{\'e}dric Pradalier$^{2}$% <-this % stops a space
  \thanks{
    $^{1}$
    UMI2958 GeorgiaTech-CNRS, 
    CentraleSup{\'e}lec, Universit{\'e} Paris-Saclay, Metz,
    Thales Group,
    {\tt\small abenbihi@georgiatech-metz.fr}
  }%
  \thanks{
    $^{2}$
    GeorgiaTech Lorraine -- UMI2958  GeorgiaTech-CNRS, Metz, France
  }%
  \thanks{
    $^{3}$
    Google Research Brain Team
  }
}

\newcommand{\mAP}{\gls{mAP}}
\newcommand{\recallN}{\textit{recall$@\!N$}}
\newcommand{\RecallN}{\textit{Recall$@\!N$}}
\newcommand{\toft}{Toft \textit{et al.}}
\newcommand{\hide}[1]{}

\definecolor{hlCellColor}{rgb}{0.95,0.95,0.95}

\makeglossaries{}
\newacronym{wrt}{\textit{w.r.t}}{with respect to}
\newacronym{mAP}{\textit{mAP}}{mean Average Precision}
\newacronym{sota}{SoA}{State-of-the-Art}
\newacronym{fig}{Fig.}{Figure}
\newacronym{tab}{Tab.}{Table}
\newacronym{bow}{BoW}{Bag of Words}
\newacronym{cnn}{CNN}{Convolutional Neural Network}
\newacronym{hog}{HoG}{Histogram of Oriented Gradients}

\begin{document}

\maketitle
\thispagestyle{empty}
\pagestyle{empty}

%%%%%%%%%%%%%%%%%%%%%%%%%%%%%%%%%%%%%%%%%%%%%%%%%%%%%%%%%%%%%%%%%%%%%%%%%%%%%%%%
\begin{abstract}
    Most of the research effort on image-based place recognition is designed
    for urban environments. In bucolic environments such as natural scenes with
    low texture and little semantic content, the main challenge is to handle
    the variations in visual appearance across time such as illumination,
    weather, vegetation state or viewpoints. The nature of the variations is
    different and this leads to a different approach to describing a bucolic
    scene. We introduce a global image description computed from its semantic
    and topological information. It is built from the wavelet transforms of the
    image's semantic edges. Matching two images is then equivalent to matching
    their semantic edge transforms. This method reaches
    state-of-the-art image retrieval performance on two multi-season
    environment-monitoring datasets: the CMU-Seasons and the Symphony Lake
    dataset. It also generalizes to urban scenes on which it is on par with
    the current baselines NetVLAD and DELF.
\end{abstract}

%%%%%%%%%%%%%%%%%%%%%%%%%%%%%%%%%%%%%%%%%%%%%%%%%%%%%%%%%%%%%%%%%%%%%%%%%%%%%%%%
\section{INTRODUCTION}

Place recognition is the process by which a place that has been observed before
can be identified when revisited. Image-based place recognition achieves this
task using images taken with similar viewpoints at different times. This is
particularly challenging for images captured in natural environments over
multiple seasons (e.g.~\cite{griffith2017symphony}
or~\cite{sattler2018benchmarking}) because their appearance is modified as a
result of weather, sun position, vegetation state in addition to view-point and
lighting, as usual in indoor or urban environments. In robotics, place
recognition is used for the loop-closure stage of most large scale SLAM systems
where its reliability is critical~\cite{cummins2011appearance}. It is also an
important part of any long-term monitoring system operating outdoor over many
seasons~\cite{griffith2017symphony,churchill2013experience}.

In practice, place recognition is usually cast as an image retrieval task where a query image is matched to
the most similar image available in a database. The search is computed on
a projection of the image content on much lower-dimensional space.
The challenge is then to compute
a compact image encoding such that images of the same location are near to each
other despite their change of appearance due to environmental changes. 

\begin{figure}[thb]
  \centering
  \includegraphics[width=\linewidth]{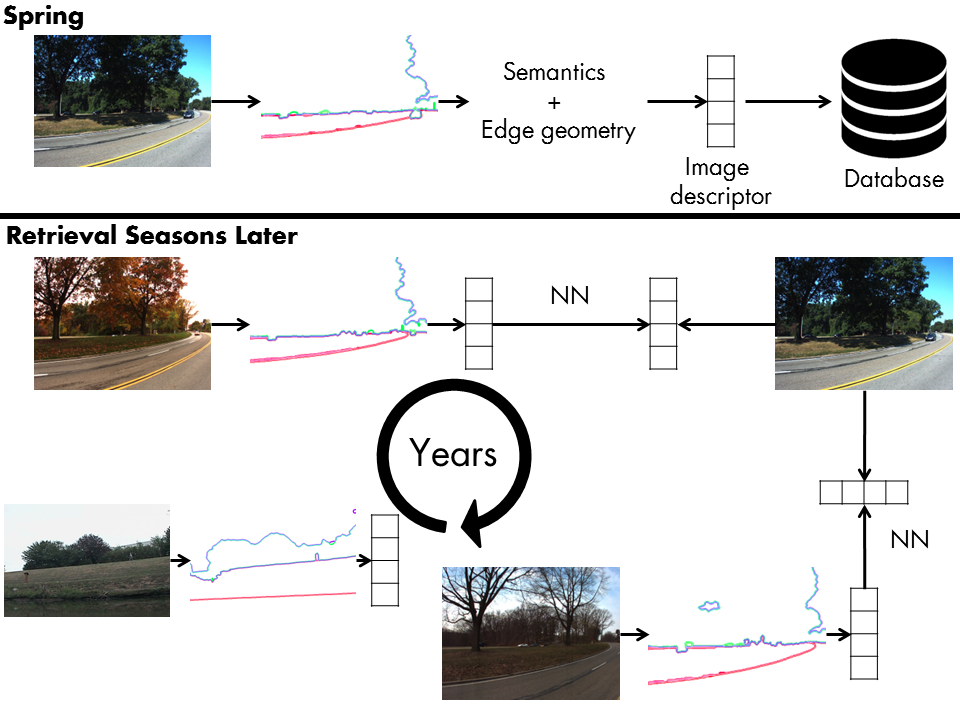}
  \caption{WASABI computes a global image descriptor for place recognition over
  bucolic environments across seasons. It builds upon the image semantics and
  its edge geometry that are robust to strong appearance variations caused by
  illumination and season changes. While existing methods are tailored for
  urban-like scenes, our approach generalizes to bucolic scenes that offer
  distinct challenges.}
  \label{fig:method}
\end{figure}

Most of the existing methods start with detecting and describing local features
over the image before aggregating them into a low-dimensional vector. The
methods differ on the local feature detection, description, and aggregation.
Most of the research efforts have focused on environments with rich semantics
such as cities or touristic landmarks~\cite{arandjelovic2016netvlad,
noh2017large}. Early methods relied on hand-crafted feature descriptions (e.g.
SIFT~\cite{lowe2004distinctive}) and simple aggregation based on histograms
constructed on a clustering of the feature space~\cite{sivic2003video}. Recent
breakthroughs use deep-learning to learn retrieval-specific detection
\cite{noh2017large}, description~\cite{babenko2014neural} and aggregation
\cite{arandjelovic2016netvlad}. Another line of work relies on the geometric
distribution of the image semantic elements to characterize it
\cite{gawel2018x}. However, all of these approaches assume that the images
have rich semantics or strong textures and focus on urban environments. On the
contrary, we are interested in scenes described by images depicting nature or
structures with few semantic or textured elements. In the following, such
environments, including lakeshores and parks, will be qualified as `bucolic'. 

In this paper, we show that an image descriptor based on the geometry of
semantic edges is discriminative enough to reach \gls{sota} image-retrieval
performance on bucolic environments. The detection step consists of extracting
semantic edges and sorting them by their label. Continuous edges are then
described with the wavelet transform~\cite{chuang1996wavelet} over a
fixed-sized subsampling of the edge. This constitutes the local description step.
The aggregation is a simple concatenation of the edge descriptors and their
labels which, together, make the global image descriptor. \gls{fig}~\ref{fig:method}
illustrates the image retrieval pipeline with our novel descriptor dubbed
WASABI\footnote{WAvelet SemAntic edge descriptor for BucolIc environment}: A
collection of images is recorded along a road during the Spring. Global image
descriptors are computed and stored in a database. Later in the year, in
Autumn, while we traverse the same road, we describe the image at the current
location. Place recognition consists of retrieving the database image which
descriptor is the nearest to the current one. To compute the image distance, we
associate each edge from one image to the nearest edge with the same semantic
label in the other image. The distance between the two edges is the
Euclidean distance between descriptors. The image distance is the sum of the distances
between edge descriptors of associated edges. 

WASABI is compared to existing image retrieval methods on two outdoor bucolic
datasets: the park slices of the CMU-Seasons\cite{sattler2018benchmarking} and
Symphony\cite{griffith2017symphony}, recorded over a period of 1 year and 3
years respectively. Experiments show that it outperforms existing methods even
when the latter are finetuned for these datasets. It is also on par with
NetVLAD, the current \gls{sota} on urban scenes, which is specifically
optimized for city environments. This shows that WASABI can generalize across
environments.

The contribution of this paper is a novel global image descriptor based on
semantics edge geometry for image-based place recognition in bucolic
environment. Experiments show that it is also suitable for urban settings. The
descriptor's and the evaluation's code are available at
\url{https://github.com/abenbihi/wasabi.git}.

\section{RELATED WORK}

This section reviews the current state-of-the-art on place recognition. A
common approach to place recognition is to perform global image descriptor
matching. The main challenge is defining a compact yet discriminative
representation of the image that also has to be robust to illumination,
viewpoint and appearance variations.

Early methods build global descriptors with statistics of local features
over a set of visual words. A first step defines a codebook of
visual words by clustering local descriptors, such as
SIFT~\cite{lowe2004distinctive}, over a training dataset. %This dataset must be
%different from the place recognition one to generalize well. 
%An image is
%described with the statistics of its local features with respect to this
%codebook. 
Given an image to describe, the \gls{bow}~\cite{sivic2003video} method assigns
each of its local features to one visual word and the output statistics are a
histogram over the words.
%are assigned to the codebook clusters and the descriptor is simply the
%clustering histogram.
The Fisher Kernels~\cite{perronnin2010large} refine this statistical model
fitting a mixture of Gaussian over the visual words and the local features.
%The
%Gaussians are learned over the training data. and computing the probability
%of each local feature . improves over the previous clustering by fitting a
%mixture of Gaussians over the training dataset local features. Then, for Each
%local feature, they It concatenates the gradient of the probability of each
%feature to belong to one of the gaussian. This high-dimensional vector is then
%reduced with Principal Component Analysis (PCA). 
This approach is simplified in VLAD~\cite{JDSP10} by
%computing clusters as in~\cite{sivic2003video} even though it does not use
%cluster-histogram for aggregation. Instead, 
concatenating the distance vector between each local feature and its nearest
cluster, which is a specific case of the derivation
in~\cite{perronnin2010large}. These methods rely on features based on pixel
distribution that assumes that images have strong textures, which is not the
case for bucolic images. They are also sensitive to variations in the image
appearance such as seasonal changes. In contrast, we rely on the geometry of
semantic elements and that proves to be robust to strong appearance changes.

Recent works aim at disentangling local features and pixel intensity through
learned feature descriptions. \cite{babenko2014neural} uses pre-trained
\gls{cnn} feature maps as local descriptors and aggregates them in a VLAD
fashion. Following work NetVLAD~\cite{arandjelovic2016netvlad} specifically
trains a \gls{cnn} to generate local feature descriptors relevant for image
retrieval. It transforms the VLAD hand-crafted aggregation into an end-to-end
learning pipeline and reaches top performances on urban scenes such as the
Pittsburg or the Tokyo time machine
datasets~\cite{torii2013visual,torii201524}. DELF~\cite{noh2017large} tackles
the problem of local feature selection and trains a network to sample only
features relevant to the image retrieval through an attention mechanism on a
landmark dataset. WASABI also relies on a \gls{cnn}s to segment images but not
to describe them. Segmentation is indeed robust to appearance changes but
bucolic environments are typically not diverse enough for the segmentation to
suffice for image description. Instead, we fuse this high-level information
with edges' geometric description to augment the discriminative power of the
description.

Similar works also leverage semantics to describe images. Toft \textit{et
al.}~\cite{toft2017long} compute semantic histograms and \gls{hog} over image
patches and concatenate them. VLASE~\cite{yu2018vlase} also relies on
semantic edges learned in an end-to-end manner~\cite{yu2017casenet}, but adopt a
description analog to VLAD. Local features are pixels that lie on a semantic
edge, and they are described with the probability distribution of that pixel to
belong to a semantic class, as provided by the last layer of the \gls{cnn}. The
rest of the description pipeline is the same as in VLAD. WASABI differs in
that it describes the geometric properties of the semantic edges and neither
semantic nor pixel statistics.

Another work~\cite{gawel2018x} that leverages geometry and semantics converts
images sampled over a trajectory into a 3D semantic graph where vertex are
semantic units and edges represent their adjacency in the images. 
%fused into a 3D graph over the database images , uses temporal information to
%fuse the graphs over time and generates a global database graph. 
A query image is then transformed into a semantic graph 
%Then, given a new image expressed as a semantic graph, 
and image retrieval is reduced to a graph matching problem. This derivation
assumes that the environment displays enough semantic to avoid ambiguous
graphs, which does not occur 
%However, this approach assumes again that the environment is rich in semantic
%elements to avoid ambiguous graphs. This is not the case 
for bucolic scenes. This is what motivates WASABI to leverage the edges'
geometry for it better discriminates between images.

%edges as another robust and discriminative image element.
The edge-based image description is not novel \cite{zhou2001edge} and the literature
offers a wide range of edge descriptors \cite{merhy2017reconnaissance}.  But
these local descriptors are usually less robust to illumination and viewpoint
variations than their pixel-based counterparts. In this work, we fuse edge
description with semantic information to reach \gls{sota} performance on
bucolic image retrieval across seasons. We rely on the wavelet descriptor for
its compact representation while offering uniqueness and invariance
properties~\cite{chuang1996wavelet}.

%%%%%%%%%%%%%%%%%%%%%%%%%%%%%%%%%%%%%%%%%%%%%%%%%%%%%%%%%%%%%%%%%%%%%%%%%%%%%%
\section{METHOD}

This section details the three steps of image retrieval: the detection and
description of local features, their aggregation, and the image distance
computation. In this paper, a local feature is constructed as a vector that
embeds the geometry of semantic edges.

\subsection{Local feature detection and extraction.} 

The local feature detection stage takes a color image as input and outputs a
list of continuous edges together with their semantic labels. Two equivalent
approaches can be considered. The first is to extract edges from the semantic
segmentation of the image, i.e. its pixel-wise classification. The \gls{sota}
relies on \gls{cnn} trained on labeled data~\cite{larsson2019cross,
zhao2017pyramid}. The second approach is also based on \gls{cnn} but directly
outputs the edges together with their labels~\cite{yu2017casenet,
acuna2019devil}. The first approach is favored as there are many more public
segmentation models than semantic edges ones.

Hence, starting from the semantic segmentation, a post-processing stage is
necessary to reduce the labeling noise. Most of this noise consists of
labeling errors around edges or small holes inside bigger semantic units. To
reduce the influence of these errors, semantic blobs smaller than
\texttt{min\_blob\_size} are merged with their nearest neighbors. 

Furthermore, to make semantic edges robust over long periods, it is
necessary to ignore classes corresponding to dynamic objects such as cars or
pedestrians. Otherwise, they would alter the semantic edges and modify the
global image descriptor. These classes are removed from the segmentation maps
and the resulting hole is filled with the nearest semantic labels.

Taking the cleaned-up semantic segmentation as input, simple Canny-based edge
detection is performed and edges smaller than \texttt{min\_edge\_size} pixels
are filtered out. 

Segmentation noise may also break continuous edges. So the remaining edges are
processed to re-connect edges belonging with each other. For
each class, if two edge extremities are below a pixel distance
\texttt{min\_neighbour\_gap}, the corresponding edges are grouped into
a unique edge.

The parameters are chosen empirically based on the segmentation noise of the
images. We use the segmentation model from~\cite{larsson2019cross}. It features
a PSP-Net~\cite{zhao2017pyramid} network trained on Cityscapes
\cite{Cordts2016Cityscapes} and later finetuned on the Extended-CMU-Seasons
dataset.
In this case, the relevant detection parameters were
\texttt{min\_blob\_size=50}, \texttt{min\_edge\_size=50} and
\texttt{min\_neighbour\_gap=5}. 

\subsection{Local feature description}

Among the many existing edge descriptor, we favor the wavelet descriptor
\cite{chuang1996wavelet} for its properties relevant to image retrieval. It
consits in projecting a signal over a basis of known function and is often used
to generate a compact yet unique representation of a signal.
Wavelet description is not the only transform to generates a unique
representation for a signal. The Fourier descriptors~\cite{zahn1972fourier,
granlund1972fourier} also provides such a unique embedding.  However, the
wavelet description is more compact than the Fourier one due to its
multiple-scale decomposition. Empirically, we confirmed that the former was
more discriminative than the latter for the same number of coefficients.
\hide{In the experiments, the wavelet transform at the first scale is already
discrminative enough for the edges. }

\begin{figure}[thb]
  \centering
  \includegraphics[width=\linewidth]{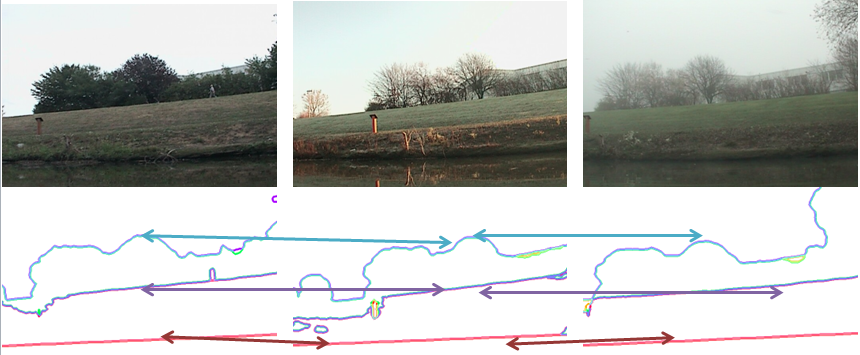}
    \caption{Symphony. Semantic edge association across strong seasonal and weather variations.}
  \label{fig:asso}
\end{figure}

Given a 2D contour extracted at the previous step, we subsample the edge at
regular steps and collect $N$ pixels. Their $(x,y)$ locations in the image are
concatenated into a 2D vector.
We compute the discrete Haar-wavelet decomposition over each axis separately
and concatenate the output that we L2 normalize.
In the experiments, we set $N=64$ and keep only the even coefficients of the
wavelet transforms. This does not destroy information as the coefficients are
redundant. The final edge descriptor is a 128-dimension vector.

\subsection{Aggregation and Image distance}

Aggregation is a simple accumulation of the edge descriptors together with
their label. Given two images and using the aggregated edge descriptors, the
image distance is the average distance between matching edges. More precisely,
edges belonging to the same semantic class are associated across the images
solving an assignment problem (see Fig.~\ref{fig:asso}). The distance used is
the Euclidean distance between edge descriptors and the image distance is the
average of the associated descriptor distances. In a retrieval setting, we
compute such a distance between the query image and every image in the database
and return the database entry with the lowest distance.

\begin{figure}[thb]
  \centering
  \includegraphics[width=\linewidth]{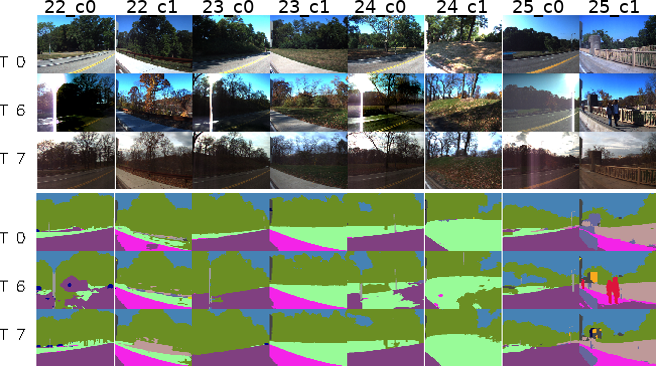}
    \caption{Extended CMU-Seasons. Top: images. Down: segmentation instead of the
    semantic edge for better visualization. Each column depicts one location
    from a \texttt{i} and a camera \texttt{j} that we note \texttt{i\_cj}.
    Each line depitcs the same location over several traversals noted
    \texttt{T}.}
  \label{fig:cmu}
\end{figure}

%%%%%%%%%%%%%%%%%%%%%%%%%%%%%%%%%%%%%%%%%%%%%%%%%%%%%%%%%%%%%%%%%%%%%%%%%%%%%%
\section{EXPERIMENTS}

This section details the experimental setup and presents results
for our approach against methods for which public code is
available: BoW\cite{sivic2003video}, VLAD\cite{jegou2011aggregating},
NetVLAD\cite{arandjelovic2016netvlad}, DELF\cite{noh2017large}, Toft \textit{et
al.}~\cite{toft2017long}, and VLASE \cite{yu2018vlase}.
We demonstrate the retrieval performance on two outdoor bucolic datasets:
CMU-Seasons\cite{sattler2018benchmarking} and 
Symphony\cite{griffith2017symphony}, recorded over a period of 1 year and 3
years respectively. Although existing methods reach SoA
performance on urban environments, our approach proves to outperform them on
bucolic scenes, and so, even when they are finetuned.
It also shows better generalization as it achieves near SoA performance of the urban
slices on the CMU-Seasons dataset.

\subsection{Datasets}

\paragraph{Extended CMU-Seasons}

The Extended CMU-Seasons dataset (Fig.~\ref{fig:cmu}) is an extended version of
the CMU-Seasons~\cite{Badino2011} dataset. It depicts urban, suburban, and park
scenes in the area of Pittsburgh, USA. Two front-facing cameras are mounted on
a car pointing to the left/right of the vehicle at approximately 45 degrees.
Eleven traversals are recorded over a period of 1 year and the images from the
two cameras do not overlap. The traversals are divided into 24 spatially
disjoint slices, with slices [2-8] for urban scenes, [9-17] for suburban and
[18-25] for the park scenes respectively. All retrieval methods are evaluated on
the park scenes for which ground-truth poses are available [22-25]. The other
park scenes [18-21] can be used to train learning approaches. For each slice in
[22-25], one traversal is used as the image database and the 10 other
traversals are the queries. In total, there are 78 image sets of roughly 200
images with ground-truth camera poses. \gls{fig}~\ref{fig:cmu} shows examples of
matching images over multiple seasons with significant variations.

\paragraph{Lake}

The Symphony~\cite{griffith2017symphony} dataset consists of 121 visual
traversals of the shore of Symphony Lake in Metz, France. The 1.3 km shore is
surveyed using a pan-tilt-zoom (PTZ) camera and a 2D LiDAR mounted on an
unmanned surface vehicle. The camera faces starboard as the boat moves along
the shore while maintaining a constant distance. The boat is deployed on
average every 10 days from Jan 6, 2014 to April 3, 2017. In comparison to the
roadway datasets, it holds a wider range of illumination and seasonal
variations and much less texture and semantic features, which challenges
existing place recognition methods. 

\begin{figure}[thb]
  \centering
  \includegraphics[width=\linewidth]{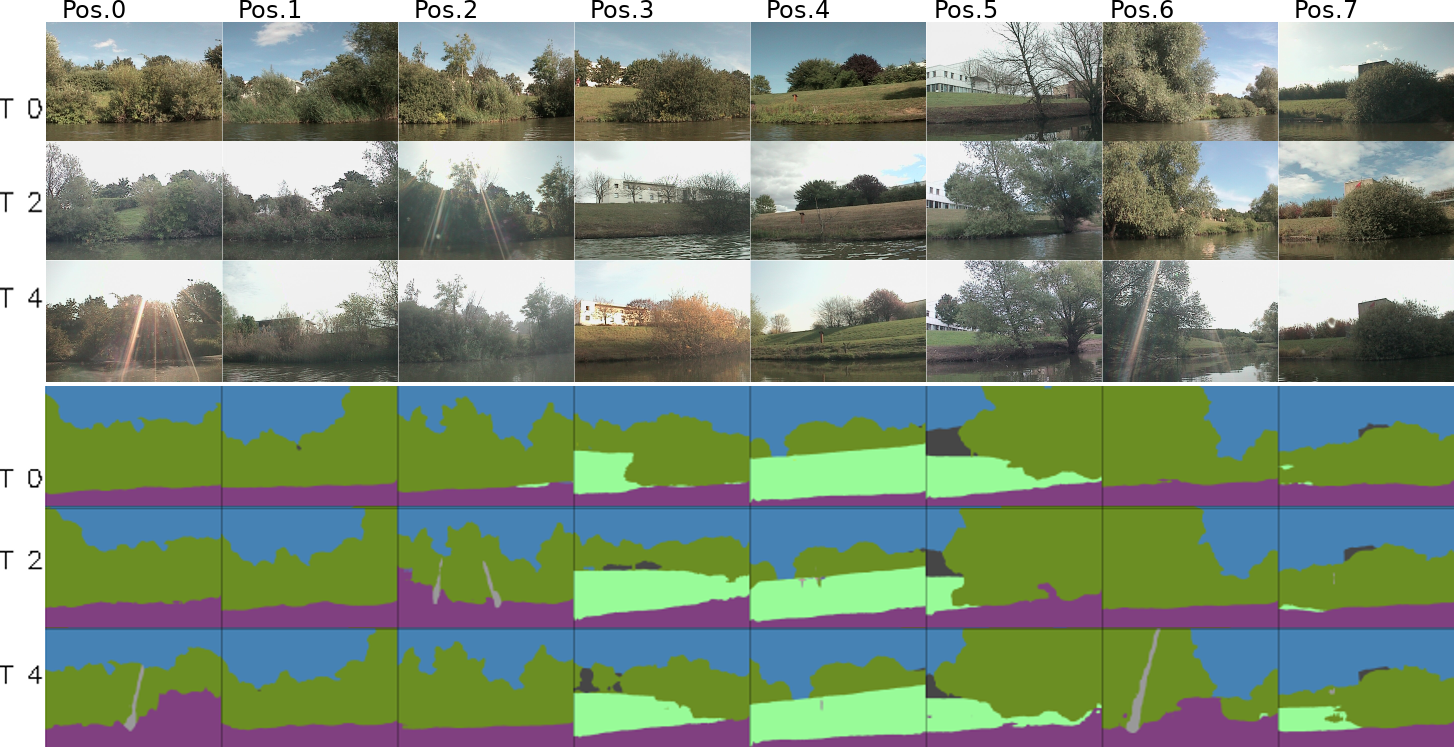}
  \caption{Symphony dataset. Top-Down: images and their segmentation. First
  line: reference traversal at several locations. Each column \texttt{k}
  depicts one location \texttt{Pos.k}. Each line depicts \texttt{Pos.k}
  over random traversals noted \texttt{T}. Note that contrary to CMU-Seasons,
  we generate mixed-conditions evaluation traversals from the actual lake
  traversals. So there is no constant illumination or seasonal condition over
  one query traversal \texttt{T}.}
  \label{fig:lake}
\end{figure}

We generate 10 traversals over one side of the lake from the ground-truth poses
computed with the recorded 2D laser scans~\cite{pradalier2018multi}. The other
side of the lake can be used for training. One of the 121 recorded traversals
is used as the reference from which we sample images at regular locations to
generate the database. For each database image, the matching images are sampled from 10
random traversals out of the 120 left. Note that contrary to the CMU-Seasons
dataset, this means that there is no light and appearance continuity over one
traversal (Fig.~\ref{fig:lake}).

\subsection{Experimental setup}

This section describes the rationale behind the evaluation. All methods are
evaluated on the CMU park and the CMU city to assess their global performance
with respect to the type of environment. The retrievals are run independently
within slices, for one camera. The experiments on the Symphony lake assess the
robustness to low texture images with few semantic elements and even harsher
lighting and seasonal variations. 
%consists of running independent place recognition over each slice and average
%the performance over the traversals. 

Also, the influence of the semantic content and the illumination is
evaluated on the CMU park.
%reOn CMU-Seasons we evaluate place recognition methods \wrt{} the semantic
%elements of a traversal on one hand, and \wrt{} the lighting and seasonal
%conditions on the other hand.
%The first CMU-Seasons evaluation, \wrt{} the semantic elements,
%consists of running independent place recognition over each slice and average the
%performance over the traversals. 
The semantic assessment is possible because each slice
%Since the slices are spatially disjoint, they
holds specific semantic elements. % that challenge the image retrieval in various ways.
For example, slice 23 seen from camera 0 holds mostly repetitive patterns of
trees that are harder to differentiate than the building skyline seen from
camera 1 on slice 25. Slices with similar semantic content are evaluated
together.
%Averaging over the traversals is a way to put aside the
%influence of the lighting and season for each traversal. 
Similarly, each traversal exhibit specific season and illumination. The
performance for one condition is computed by averaging the retrieval
performance of traversals with similar conditions over all slices.
%The second evaluation, \wrt{} the lighting and season, starts the
%same way with independent place recognition over each slice. But the scores are
%averaged over the slices for each traversal. This way, the semantic content is
%the same for all the traversals and only the lighting and season change.

As mentioned previously, our approach is evaluated against BoW, VLAD, NetVLAD,
DELF, \toft, and VLASE. In their version available online, these methods are
mostly tailored for rich urban semantic environments: the codebook for BoW and
VLAD is trained on Flickr60k~\cite{jegou2008hamming}, NetVLAD is trained on the
Pittsburg dataset \cite{torii2013visual}, DELF on the Google landmark
one~\cite{noh2017large}, and VLASE on the CaseNet model trained on
are the Cityscape dataset \cite{cordts2016cityscapes}. For fair comparison, we
finetune them on CMU-Seasons and Symphony when possible, and report both
original scores and the finetuned ones noted with (*). 

A new 64-words codebook is generated for BOW and VLAD, using the CMU park images
from slices 18-21. The NetVLAD training requires images with ground-truth
poses, which is not the cases for these slices. So it is trained on three
slices from 22-25 and evaluated on the remaining one. On Symphony, images
together with their ground-truth poses are sampled from the east side of the
lake that is spatially disjoint from the evaluation traversals. The DELF local
features are not finetuned as the training code is not available even though
the model is. The segmentation model~\cite{larsson2019cross} used for \toft's
descriptor is the same as for WASABI and was trained to segment the CMU park
across seasons. The CaseNet model used by VLASE is not finetuned.

\toft descriptor is the concatenation of semantic and pixel oriented gradients
histograms over image patches. In the paper, \toft divide the top half of the
image 6 rectangle patches ($2 \times 3$ lines-column). The \gls{hog} is
computed by 
%To compute the \gls{hog}, each patch is 
further dividing into smaller rectangles and concatenating each of their
histograms.
%over which the gradient orientations are accumulated. 
The descriptor performance depends on the resolution of these two splits and 
%\gls{hog}'s resolution and the one at which the patches are sampled. Also, 
its dimension increases with it. A grid of parameters is tested and the best
results are reached when it is the highest: this is expected as such a
resolution embeds more detail about local image information which helps
discriminate between image. The 7506-dim descriptor is derived from the top
two-thirds of the image divided into $6\times9$ patches, with into $4 \times 4$
sub-rectangles and the \gls{hog} has an 8-bin discretization. Further
dimensionality reduction is out of the scope of this paper, and we report the
results for both the high-resolution and the original one.
%However, the descriptor memory overhead increases with the resolution.  We
%report results for both the paper's resolution and the highest computed
%resolution for which the descriptor is in $\mathbb{R}^{7506}$.

A grid search is also run on VLASE to select the probability threshold $T_e$ above
which a pixel is a local feature, and the maximum number of features to keep in
one image. The best results are reported for $T_e=0.5$ and 3000 features per
image.

%Finally, our approach is tested against the original available methods on the
%three urban CMU-Seasons slices for which ground-truth poses are
%available~[6-8]. This assesses whether our approach is also relevant for urban
%settings and hence better generalize across environments than methods tailored
%specifically for urban scenes.

\subsection{Metrics}

The place recognition metrics are the \recallN{} 
and the \mAP{}\cite{philbin2007object}.
Both depend on a distance threshold $\epsilon$: a retrieved database image matches
the query if the distance between their camera center is below $\epsilon$.
The \recallN{} is the percentage of queries for which there is at least one
matching database image in the first $N$ retrieved images. We set $N \in
\{1,5,10,20\}$, and $\epsilon$ to $5m$ and $2m$ for the CMU-Seasons and the
Symphony datasets respectively. Both metrics are available in the code.

\subsection{Results}

\begin{figure}[thb]
  \centering
  \includegraphics[width=\linewidth]{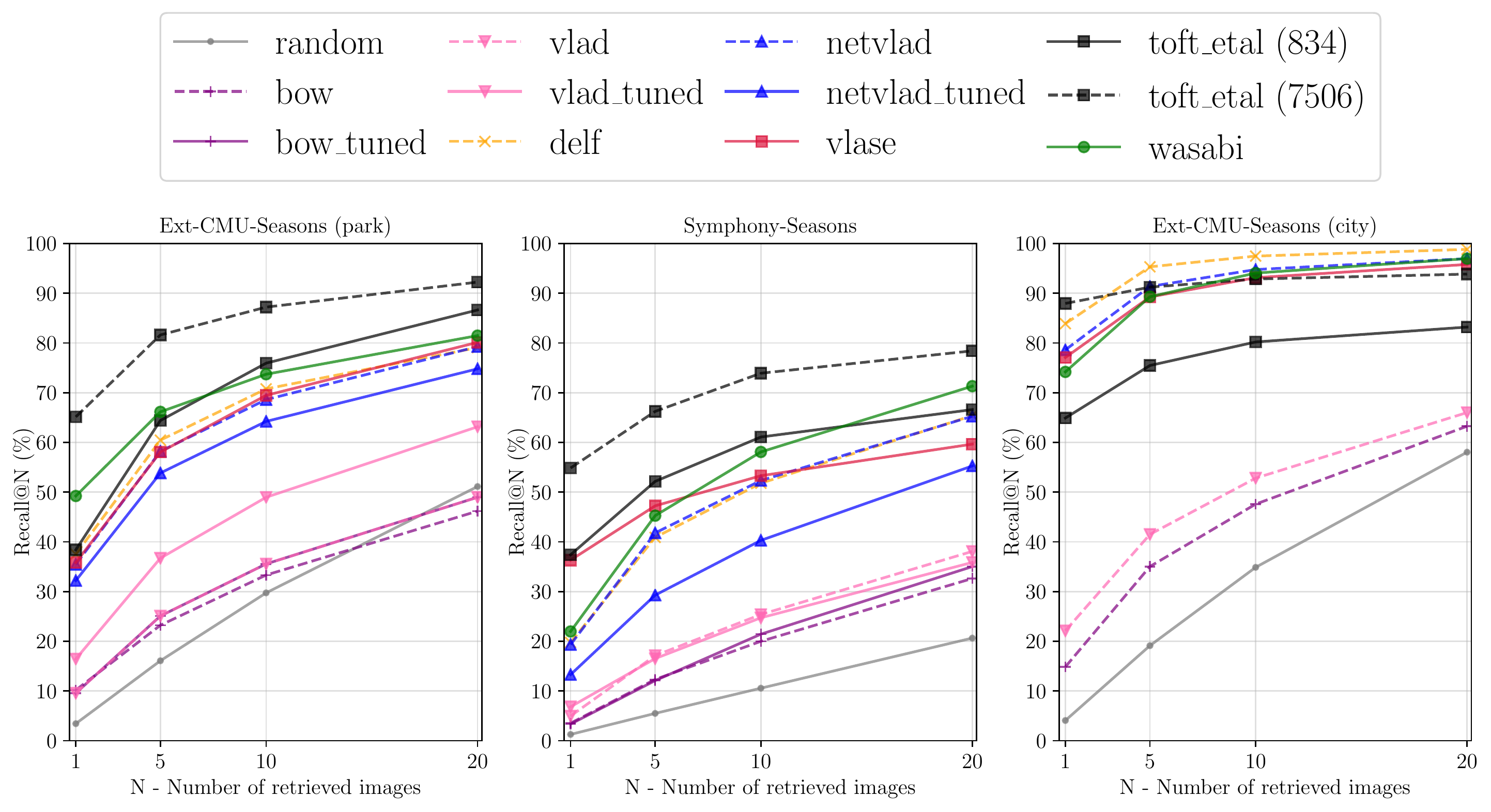}
  \caption{Retrieval performance for each dataset measured with the \RecallN{}.
  Retrieval is performed based on the similarity of the descriptors and no
  further post-processing is run for all methods. The high-resolution
  description from \cite{toft2017long} reaches the best score, followed by
  WASABI and current \gls{sota} methods. These results suggest that a
  hand-designed descriptor can compare with existing deep approaches.}
  \label{fig:perf_global}
\end{figure}

\begin{figure}[thb]
  \centering
  \includegraphics[width=\linewidth]{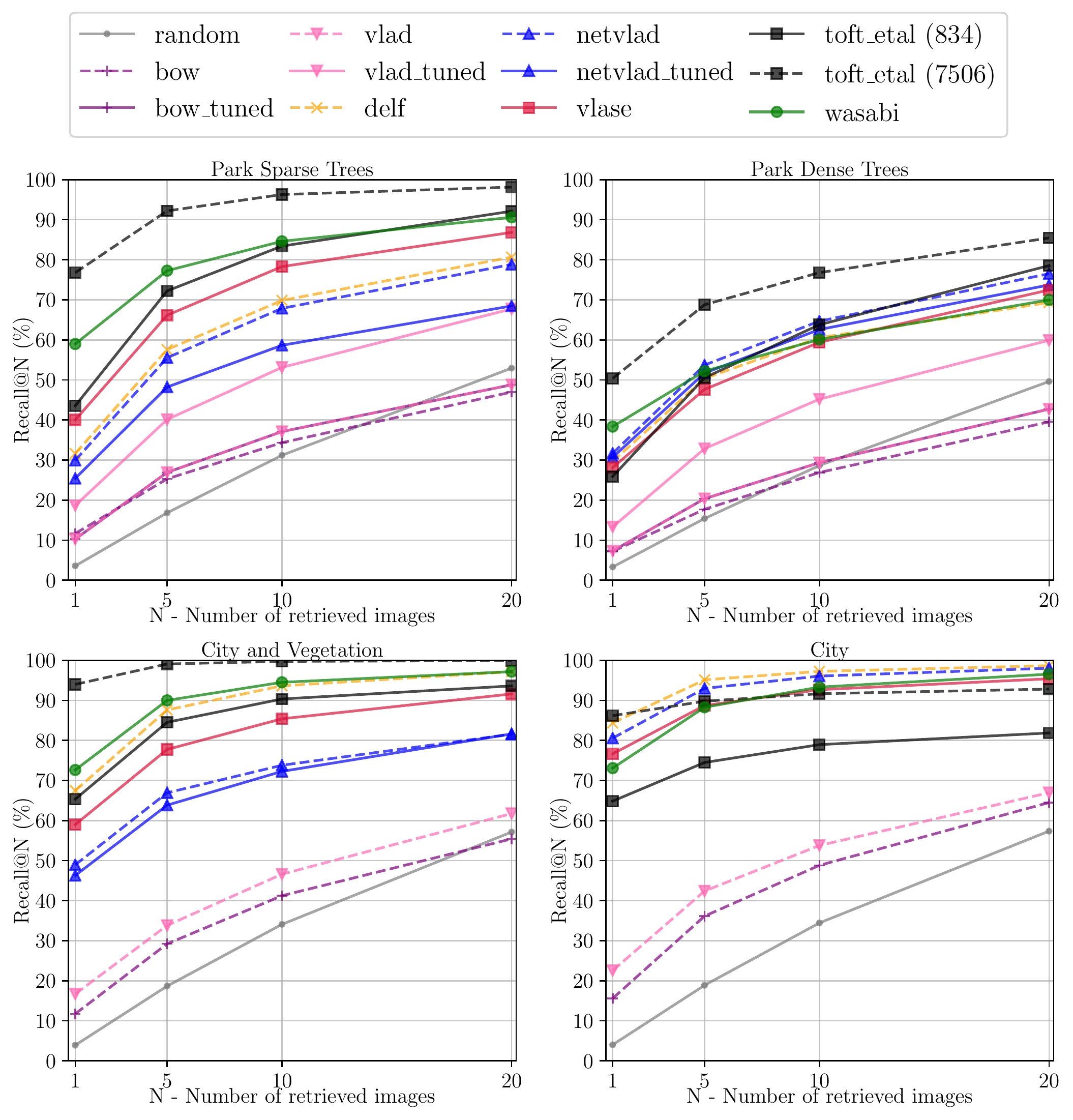}
  \caption{Ext-CMU-Seasons. Retrieval results on urban scenes with vegetation
  elements \textit{v.s.} urban scenes with only city structures.}
  \label{fig:semantics}
\end{figure}

Overall, semantic-based methods are better suited than existing ones to
describe bucolic scenes, even when they are originally tailored for urban
environments such as VLASE and~\cite{toft2017long}. WASABI achieves better
performance when the semantic segmentation is reliable (CMU-Park) but is less
robust when it exhibits noise (Symphony) (\gls{fig}~\ref{fig:perf_global}). It
generalizes well to cities and appears to better handle the vegetation
distractors than \gls{sota} methods (\gls{fig}~\ref{fig:semantics}) The rest of
this section details the results. 

\paragraph{Global Performance}

%Experiments suggest that approaches based on the fusion of semantics and
%hand-design, such as the WASABI descriptor and~\cite{toft2017long,yu2018vlase},
%are as relevant for scene recognition as end-to-end deep
%approaches~\cite{arandjelovic2017netvlad, noh2017large}. 

\gls{fig}~\ref{fig:perf_global} plots the \RecallN{} over the three types of
environments: the CMU park, the Symphony lake, and the CMU city. Overall, the
method from Toft \textit{et al.}~\cite{toft2017long} achieves the best results
when it aggregates local descriptors at a high resolution (toft\_etal~(7506)).
The necessary memory overhead may be addressed with dimensionality reduction
but this is out of the scope of this paper.

WASABI achieves the $2^{nd}$ best performance of the CMU park while only
slightly underperforming the \gls{sota} NetVLAD and DELF on urban environments.
This is expected as the \gls{sota} is optimized for such settings. Still, this
shows that our method generalizes to both types of environments. 
Also, this suggests that semantic edges are discriminative enough to recognize
a scene, even when there are few semantic elements such as in the park.
This assumption is comforted by the satisfying performance of VLASE, which also
leverages semantic edges to compute local features. Note that while it
underperforms WASABI on the CMU-Park, it provides better results on the
Symphony data, as does~\cite{toft2017long}. 

\begin{figure}[thb]
  \centering
  \includegraphics[width=0.8\linewidth]{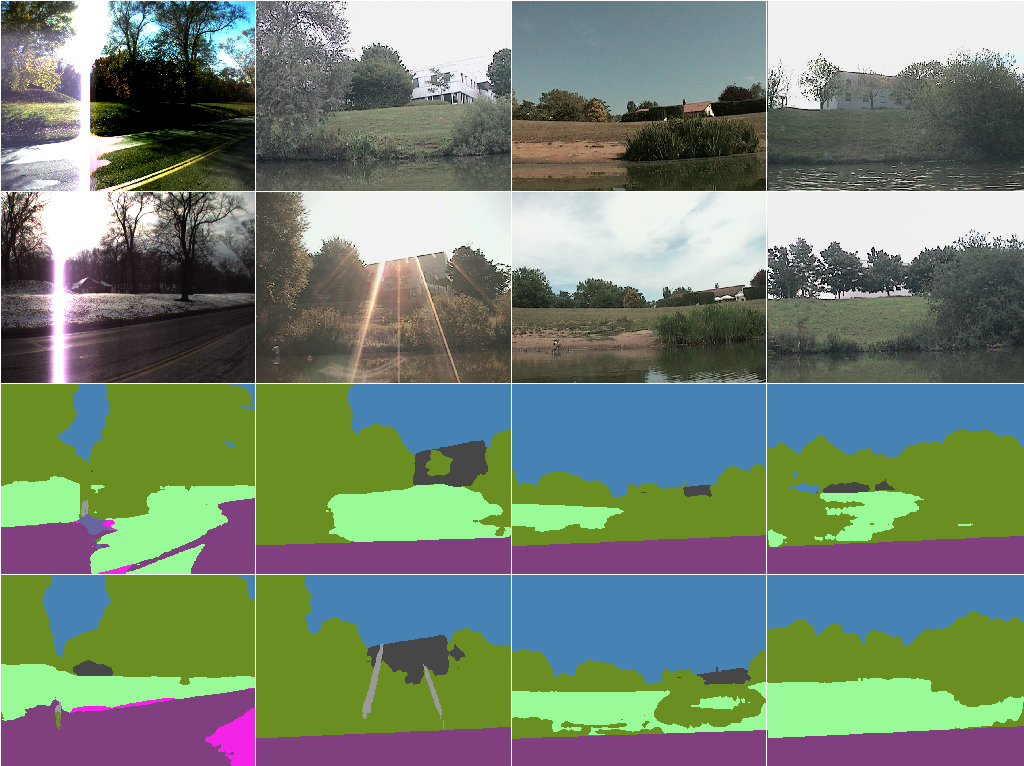}
  \caption{Segmentation failures. Left column: CMU-Seasons. A strong sunglare is
  present along survey 6 (sunny spring) or 8(snowy winter). Other columns:
  Symphony. The segmentation is not finetuned on the lake and produces a noisier output. It is also sensitive to sunglare.}
  \label{fig:sunglare}
\end{figure}

On Symphony, WASABI falls behind VLASE and Toft's descriptor. This is
unexpected given the satisfying results on the CMU-Park.
%One explanation is the segmentation noise induced by the image
%noise in one hand (\textit{e.g.} sunglare) and the lack of domain adaptation on
%the other hand (Fig.~\ref{fig:sunglare}). As there is no ground-truth
%segmentation for the Symphony dataset, finetuning the segmentation is currently
%not possible. 
%
%However, the satisfying results on CMU-Seasons motivate future
%work to improve the Symphony segmentation as well as the robustness of the
%descriptor to failures of the segmentation stage.
There are two main explanations for these poor results: the
first is that the segmentation model trained for the CMU images generates noisy
outputs on the Symphony images, especially around the edges
(\gls{fig}~\ref{fig:icra20_sun_glare}). So the WASABI
wavelet descriptors can not be consistent enough across images. One reason that
allows~\cite{toft2017long} and VLASE to be robust to this noise is that they do
not rely on the semantic edge geometry directly: Toft \textit{et al.} leverages
the semantic information in the form of a label histogram which is less
sensitive to noise than segmentation itself. A similar could explain VLASE's
robustness even tough it samples local features from those same semantic edges.
The final histogram of semantic local feature is less sensitive to semantic
noise than the semantic edge coordinates on which WASABI relies.
The second explanation for WASABI's underperformance on WASABI is the smaller
edge densities compared to the CMU data. This suggests that the geometric
information should be leveraged at a finer scale than the edge's one. This is
addressed in the next chapter along with the scalability issues.

Finetuning existing methods to the bucolic scenes proves to be useful for VLAD
only on CMU-Park but does not improve the overall performance for BoW and
NetVLAD. A plausible explanation is that these methods require more data than
the one available. Investigating the finetuning of these methods is out of the
scope of this paper.

\paragraph{Semantic analysis}

\gls{fig}~\ref{fig:semantics} plots the retrieval performance on CMU scenes
with various semantics. Overall, WASABI exhibits a significant advantage over
\gls{sota} on scenes with sparse bucolic elements (top-left). However,
%It is limited by the amount
%of geometric information available: 
%when the images depicts dense foliage, the
%performance drops to the level of existing methods. 
%It compares to NetVLAD and
%DELF although it is not specifically tailored for it. 
%This shows that although
%hand-designed, the WASABI visual features can generalize.
When the slices hold mostly dense trees along the road, all performances drop
because the images hold repetitive pixel intensity patterns, few edges and
little semantic information. This limits the amount of information WASABI can
rely on to summarize an image, which restrains the description space. 
As for VLASE, in addition to the few edges to leverage, the highly repetitive
patterns lead to similar semantic edge probabilities. So their aggregation into
an image descriptor is not discriminative enough to differentiate such scenes.
A similar explanation holds for the \cite{toft2017long} descriptor of which
both the semantic histogram and SIFT-like descriptors are similar across
images. Once again, the main cause is the redundancy of the image information.

Note that this is also an open problem for urban environments. One of the
few works that tackle this specific problem is~\cite{torii2013visual}. Torii
\textit{et al.} propose to weight the aggregation of local features so that
repetitive ones do not dominate the sparser one. However, this processing can
not be integrated as is since dense bucolic scenes are entirely dominated by
redundant patterns. So there is no other discriminative information to balance
them with. The integration of such balancing is the object of future work to
tackle the challenge of repetitive patterns in natural scenes.

\paragraph{Illumination Analysis}

\begin{figure}[thb]
  \centering
  \includegraphics[width=\linewidth]{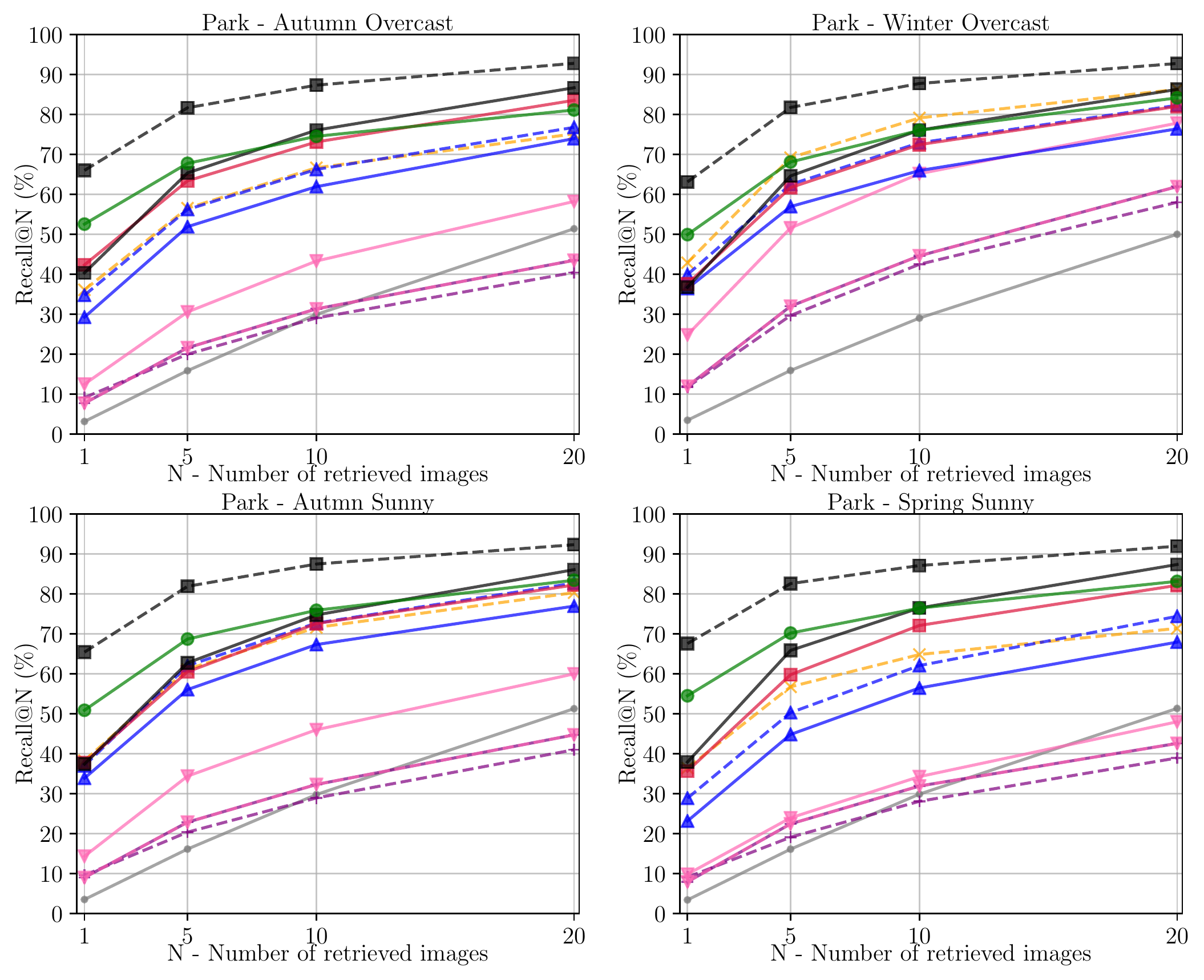}
  \caption{Ext-CMU-Seasons. Retrieval results on urban scenes with vegetation
  elements \textit{v.s.} urban scenes with only city structures.}
  \label{fig:light}
\end{figure}

\gls{fig}~\ref{fig:light} plots the retrieval performance on the CMU Park for
various illumination conditions, knowing that the reference traversal was sampled
during a sunny winter day. The query traversals with sunglare (\textit{e.g} the
sunny winter one) have been removed to investigate the relation only between the
light and the scores. 

There seems to be no independent correlation between the query's illumination
and the results, with less than 4\% variance across the conditions. This
suggests that intra-season recognition, such as between a sunny winter day and
an overcast one, is as hard as with a sunny spring day. Intriguingly, handling
pixel domain variations may be as challenging as content variations. One
explanation may be that illumination variations can affect the image as much as
a content change. For example, both light and season can affect the leaves
color. However, this does not justify the average performance of the winter
compared to other seasons even though the reference traversal was sampled in
spring.

%\vspace{-3mm}
\section{CONCLUSIONS}

In this paper, we presented WASABI, a novel image descriptor for place
recognition across seasons in bucolic environments. It represents the image
content through the geometry of its semantic edges, taking advantage of their
invariance \gls{wrt} seasons, weather and illumination. Experiments show that
it is more relevant than existing image-retrieval approaches whenever the
segmentation is reliable enough. It even generalizes well to urban settings on
which it reaches scores on par with \gls{sota}. However, it does not handle
the segmentation noise as well as other methods.
%better suited for urban environments. Tuning these methods for bucolic
%datasets proves to be insufficient to reach the same performances as our
%approach. Conversely, 
Current research now focuses on disentangling the image description from the noise
segmentation.

\addtolength{\textheight}{-1cm}   % This command serves to balance the column lengths
                                  % the textheight of the last page by a suitable* amount.

%%%%%%%%%%%%%%%%%%%%%%%%%%%%%%%%%%%%%%%%%%%%%%%%%%%%%%%%%%%%%%%%%%%%%%%%%%%%%%%%

\bibliography{root}
\bibliographystyle{IEEEtran}

\end{document}